\pdfoutput=1

\documentclass[11pt]{article}

\usepackage{amsmath,amsfonts,bm}

\def\eqref#1{equation~\ref{#1}}

\def\1{\bm{1}}

\DeclareMathAlphabet{\mathsfit}{\encodingdefault}{\sfdefault}{m}{sl}
\SetMathAlphabet{\mathsfit}{bold}{\encodingdefault}{\sfdefault}{bx}{n}

\def\sV{{\mathbb{V}}}

\usepackage[]{acl}

\usepackage{times}
\usepackage{latexsym}

\usepackage[T1]{fontenc}
\usepackage{cleveref}

\usepackage[utf8]{inputenc}

\usepackage{microtype}
\usepackage{soul}

\newcommand{\wyshi}[1]{\textcolor{black}{#1}}

\usepackage{multirow}
\usepackage[T1]{fontenc}
\usepackage{algorithm}
\usepackage{algorithmic}

\usepackage[export]{adjustbox}
\usepackage{makecell}
\usepackage{booktabs}

\usepackage{float}
\usepackage{color,soul}

\usepackage{colortbl}
\usepackage{amsmath}
\usepackage{xspace}
\definecolor{mygray}{gray}{.9}

\usepackage{amssymb}
\usepackage{pifont}
\usepackage{xcolor}

\usepackage{algorithm}
\usepackage{algorithmic}
\usepackage{microtype}
\usepackage{amsmath}
\usepackage{multirow}
\usepackage{amsmath,array,graphicx,amssymb}
\usepackage[export]{adjustbox}
\usepackage{subcaption}
\usepackage{multirow}
\usepackage{mathtools} 
\usepackage{tabularx}
\usepackage{breakcites}
\usepackage{xcolor}
\usepackage{array}
\usepackage{makecell}
\usepackage{tabularx}
\usepackage{mathrsfs}
\usepackage{scalerel}
\usepackage{soul}
\usepackage{colortbl}
\definecolor{mygray}{gray}{.95}
\usepackage{soul}
\usepackage{booktabs}

\usepackage{graphicx}
\usepackage{caption}
\usepackage{algorithm}
\usepackage{algorithmic}
\usepackage{microtype}
\usepackage{amsmath}
\usepackage{multirow}
\usepackage{amsmath,array,graphicx,amssymb}
\usepackage{caption}

\usepackage{amsthm}

\newcommand\eg{e.g.,\xspace}
\newcommand\ie{i.e.,\xspace}

\newcommand{\nclassifier}{\ensuremath{N_R}\xspace}
\usepackage{soul}
\usepackage{color}
\usepackage{xcolor}
\usepackage{enumitem}

\newcommand\blfootnote[1]{%
  \begingroup
  \renewcommand\thefootnote{}\footnote{#1}%
  \addtocounter{footnote}{-1}%
  \endgroup
}

\definecolor{apricot}{RGB}{255,240,234}
\definecolor{amber}{RGB}{214,81,30}

\definecolor{lightcyan}{RGB}{229,250,245}
\definecolor{teal}{RGB}{0,121,86}

\definecolor{darkbrown}{HTML}{784212}
\definecolor{darkgray}{HTML}{424949}
\definecolor{darkred}{HTML}{922B21}
\definecolor{darkpurple}{HTML}{5B2C6F}
\definecolor{darkblue}{HTML}{1A5276}
\definecolor{darkgreen}{HTML}{196F3D }
\definecolor{darkkhaki}{rgb}{0.74, 0.72, 0.42}
\definecolor{darkorange}{rgb}{1.0, 0.55, 0.0}

\usepackage{xspace}

\newcommand{\england}{\textbf{\textcolor{darkpurple}{\textsc{England}}}\xspace}
\newcommand{\russia}{\textbf{\textcolor{darkgray}{\textsc{Russia}}}\xspace}
\newcommand{\austria}{\textbf{\textcolor{darkred}{\textsc{Austria}}}\xspace}
\newcommand{\italy}{\textbf{\textcolor{darkgreen}{\textsc{Italy}}}\xspace}
\newcommand{\germany}{\textbf{\textcolor{darkbrown}{\textsc{Germany}}}\xspace}
\newcommand{\france}{\textbf{\textcolor{darkblue}{\textsc{France}}}\xspace}
\definecolor{candypink}{rgb}{0.89, 0.44, 0.48}
\definecolor{debianred}{rgb}{0.84, 0.04, 0.33}
\definecolor{ourlightblue}{HTML}{E0ECF7}
\definecolor{ourdarkblue}{HTML}{092E6B}
\definecolor{msgrblue}{HTML}{4889f4}
\definecolor{msgrgray}{HTML}{e1e1e7}
\definecolor{msgrpaleblue}{HTML}{a9c6f5}
\definecolor{palegreen}{HTML}{c0eeC3}
\definecolor{palepurple}{HTML}{e5d1f8}
\definecolor{paleorange}{HTML}{f9dbb1}
\definecolor{palered}{HTML}{F5B7B1}
\definecolor{jonquil}{rgb}{0.98, 0.85, 0.37}

\definecolor{myblue}{HTML}{a4ceda}
\definecolor{mygray}{HTML}{dee6e6}
\definecolor{mygreen}{HTML}{dae3d2}
\definecolor{myred}{HTML}{c04228}

\newcommand{\contexta}[1]{{\colorbox{myblue}{\parbox{19em}{#1}}}}
\newcommand{\contextb}[1]{{\colorbox{mygray}{\parbox{19em}{#1}}}}

\newcommand{\botb}[1]{{\colorbox{mygreen}{\parbox{19em}{#1}}}}

\definecolor{humanc}{rgb}{0.8, 0.8, 0.8}

\newcommand*{\myalign}[2]{\multicolumn{1}{#1}{#2}}

\usepackage{tikz}

\def \DataName{DiplomacyNonsense\xspace}
\def \ModelName{\textsc{AutoReply}\xspace}

\title{\ModelName: Detecting Nonsense in Dialogue Introspectively \\ with Discriminative Replies

}

\author{First Author \\
  Affiliation / Address line 1 \\
  Affiliation / Address line 2 \\
  Affiliation / Address line 3 \\
  \texttt{email@domain} \\\And
  Second Author \\
  Affiliation / Address line 1 \\
  Affiliation / Address line 2 \\
  Affiliation / Address line 3 \\
  \texttt{email@domain} \\}

\author{Weiyan Shi$^{1,2\dagger}$, Emily Dinan$^{2}$, Adi Renduchintala$^{2\diamond}$, \\ \textbf{Daniel Fried}$^{2,3\ddagger}$, \textbf{Athul Paul Jacob}$^{2,4}$, \textbf{Zhou Yu}$^1$, \textbf{Mike Lewis}$^2$\\
Columbia University$^1$, Meta AI$^2$, Carnegie Mellon$^3$, MIT$^4$ \\
\texttt{\{ws2634,zy2461\}@columbia.edu}, \texttt{\{edinan,mikelewis\}@meta.com } \\
\texttt{adithyare@nvidia.com}, 
\texttt{dfried@cs.cmu.edu}, 
\texttt{apjacob@mit.edu}}

\begin{document}
\maketitle
\begin{abstract}
\wyshi{Existing approaches built separate classifiers to detect nonsense in dialogues. In this paper, }
we show that without external classifiers, dialogue models can detect errors in their own messages introspectively, by calculating the likelihood of replies that are indicative of poor messages. For example, if an agent believes its partner is likely to respond \emph{``I don’t understand''} to a candidate message, that message may not make sense, so an alternative message should be chosen. We evaluate our approach on a dataset from the game Diplomacy, which contains long dialogues richly grounded in the game state, on which existing models make many errors. We first show that hand-crafted replies can be effective for the task of detecting nonsense in applications as complex as Diplomacy. We then design \ModelName, an algorithm  to search for such discriminative replies automatically, given a small number of annotated dialogue examples. We find that \ModelName-generated replies outperform handcrafted replies and perform on par with carefully fine-tuned large supervised models. \wyshi{Results also show that one single reply  without much computation overheads can also detect dialogue nonsense reasonably well. \blfootnote{$\dagger$ Work done when interning at Meta AI.} \blfootnote{$\diamond$ Work done at Meta AI, now at Nvidia.} \blfootnote{$\ddagger$ Work done during postdoc at Meta AI, now at CMU.}    }

\end{abstract}

\section{Introduction}

Detecting nonsensical dialogue generation has been an enduring challenge in dialogue research \cite{welleck2018dialogue, shi2020refine, nie2020like}. 
Previous work proposed datasets with bad message annotations  and built classification models to detect them. 
But such a supervised learning approach \wyshi{requires building an extra model without fully utilizing the dialogue model's own language ability. } 
In complex dialogue applications, data annotation is often limited but the space of bad messages is  large, ranging from easy (\eg repetition), grounding-related (\eg contradiction to context)  to challenging even for human novices (\eg domain-specific mistakes). 
In this paper, we refer to all the mistakes made by the dialogue model as nonsense. 
Previously, \citet{mehri2020unsupervised} combined the probabilities of hand-crafted replies like \emph{``that's different from what you said''}, \emph{``Wow! Very interesting.''}  to get a score for dialogue quality evaluation. Inspired by this, we utilize the dialogue model itself and  develop a reply-based approach that uses the probability of discriminative follow-up replies to detect if a message is nonsensical. 
Intuitively, if a message doesn't make sense like in Table~\ref{fig:intro dial examples}, the probability of a follow-up reply like \emph{``I don't understand your message''} will be high. 


\begin{table}[!t]
\centering
\small
\begin{tabular}{p{22em}}
\toprule
\textbf{Nonsensical message and follow-up reply} \\
\midrule
\myalign{l}{\contextb{\france: i won't move out of spain, just don't move out of marseilles}}\\
\myalign{r}{\contexta{\italy: Thank you. You will be watching my moves.}} \\
\myalign{l}{\contextb{\france: ok}}\\
\myalign{r}{\contexta{\italy: :)}} \\
\myalign{l}{\contextb{\france: \textcolor{myred}{\textbf{are you moving to spain, spain?}}}} \\
\midrule
\myalign{r}{\botb{\italy: I don't understand your message.}} \\

\bottomrule
\end{tabular}
    \caption{A dialogue in the Diplomacy game, where the players for France and Italy negotiate their actions around moving into Spain or not. In the message in bold in red, France mistakes the partner as Spain (the partner should be Italy). So the message is nonsensical, and the probability of it followed by a reply like \emph{``I don't understand your message''} is high. Our goal is to automatically generate many such follow-up replies.}
\vspace{-1em}
\label{fig:intro dial examples}
\end{table}

However, like prompt engineering, manually designing such replies is labor intensive and hard to scale to broad classes of errors. 
Therefore, we propose \ModelName, which uses the dialogue model itself and a small set of nonsense-annotated dialogue examples, and does not require hand-crafting replies or training an external supervised classifier. \ModelName is a contrastively-pruned search procedure that finds discriminative replies by iteratively selecting tokens that are highly likely after multiple nonsensical messages. The motivation is that, if a given token is highly likely after multiple nonsensical messages, then it likely captures nonsense-related features and may lead to discriminative replies. We  enumerate and expand such tokens to generate replies, and  then contrast the generated replies against good messages to prune the search and keep the discriminative replies only. 

We evaluate our reply-based approach on dialogue models for the board game Diplomacy \cite{calhamer1959diplomacy}. Diplomacy involves lengthy dialogues in which players attempt to persuade others to take certain actions in the game. This setting represents a challenge to existing dialogue models, as messages often contain detailed discussions of both past events and hypothetical future actions in the game, providing many opportunities for subtle errors in messages. Even large pre-trained models, and non-expert humans, make frequent mistakes in messages---and discriminating good from bad messages often requires complex reasoning. 



To summarize, our contributions are three-fold. First, we propose to utilize the dialogue model itself and discriminative replies to detect the models' own mistakes  and demonstrate that such an introspective reply-based approach is effective  in Diplomacy, a complex dialogue setting. 
Second, to reduce the manual work in reply design, we propose \ModelName to automatically generate a large number of discriminative replies from limited annotations. 
Third, experiments show that \ModelName can automatically generate many high-quality discriminative replies, and  these replies achieve substantially better results than hand-crafted replies, and perform on par with large fine-tuned supervised models without actually training extra classifiers.  

We note that our main goal is not to build state-of-the-art nonsense detector but rather, to explore a drastically new approach to dialogue nonsense detection, which utilizes the dialogue model's own ability to introspect, roll out to the future, and detect its own mistakes without  external classifiers or new parameters. We show promising results and hope to inspire more research in similar directions.  



\section{Methods}

\begin{figure*}
    \centering
    \includegraphics[scale=0.6]{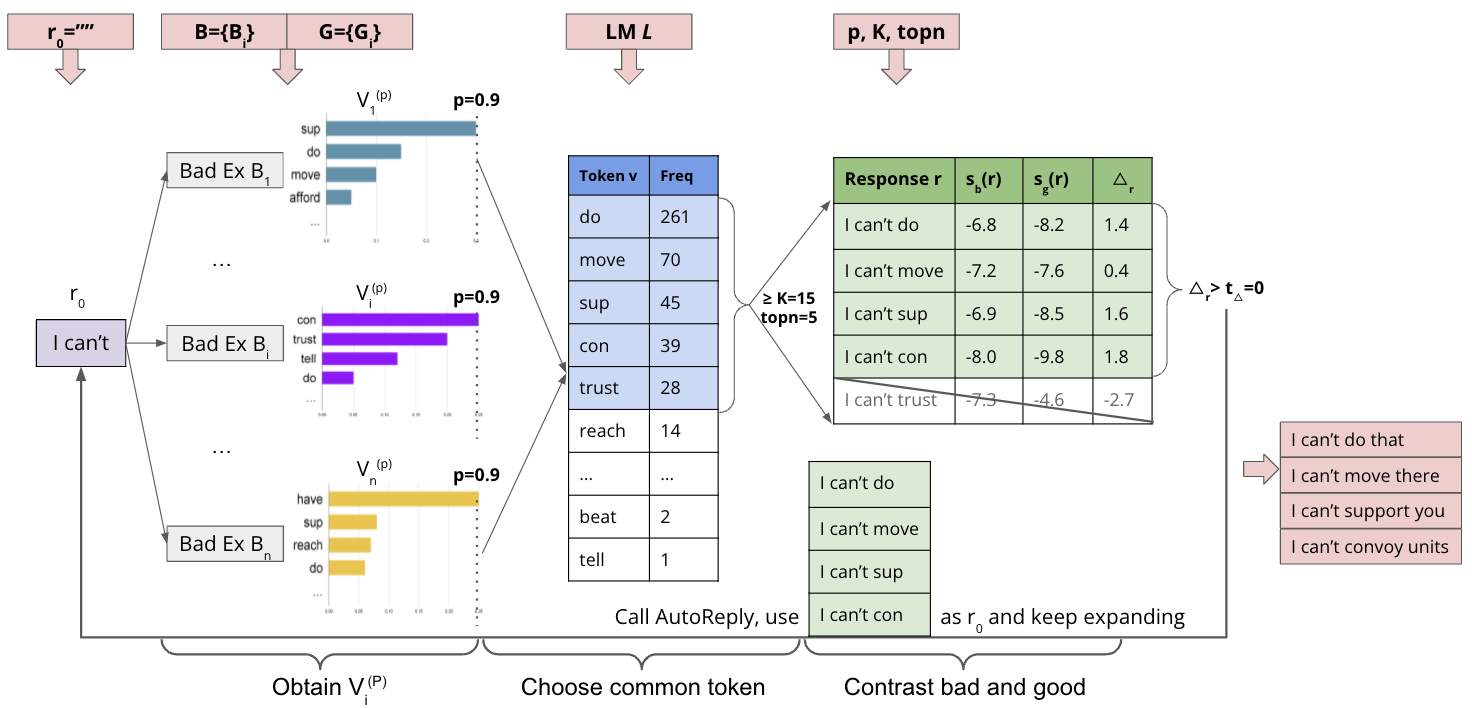}
    \caption{ \ModelName generates replies recursively. Given a response prefix $r_0$, for each bad example $B_i$, we find a high-probability token set $V_i^{(p)}$. Then we aggregate $\{\sV_i^{(p)}\}$, get each token's count and select the $topn$ tokens with  count $\geq K$. Next we contrast the probability of the generated reply $r=r_0+v$ under bad examples against the probability under good examples and select those tokens with $\Delta_r>t_{\Delta}$ to further expand in the next recursive step. 
    }
    \label{fig:model}
\end{figure*}



We propose to detect whether a message $x$ from a dialogue language model $\mathcal{L}$ is nonsensical by using the model's own distribution over possible reply messages $r$: $P_{\mathcal{L}}(r|x)$.\footnote{The models we use also condition on other elements of the context, \eg the previous messages and board state, but we omit these from the notation for brevity.
Our \ModelName search method assumes that $\mathcal{L}$ produces replies incrementally, token-by-token, but places no other restrictions on the dialogue model. 
In our experiments, we use fine-tuned BART sequence-to-sequences models for $\mathcal{L}$, see Section~\ref{sec:experiments}.
}
We first describe  how to detect nonsense with a set of hand-crafted replies in Section~\ref{sec:baseline}, and then describe  how to automatically generate replies with \ModelName in Section~\ref{sec:autoreply}. 

\subsection{Hand-Crafting Replies}
\label{sec:baseline}
For a baseline approach using hand-crafted replies, we first have human experts analyze messages generated by the dialogue model  to categorize types of nonsense and then carefully  design suitable replies for each type, e.g., \emph{``I don't understand''} for general nonsense, or \emph{``I don't have any units to do that''} for proposing an invalid move.  See Table~\ref{tab:hand-crafted replies} in the Appendix for a full list of hand-crafted replies,

We use each hand-crafted reply $r$ together with the dialogue model $\mathcal{L}$  to construct a threshold-based nonsense classifier: for a given example $x$, if $P_\mathcal{L}(r|x)> t_r$,  we predict $x$ as nonsense, using a reply-specific parameter $t_r$. 
We ensemble together the resulting classifiers using a voting-based ensembling scheme: a message is classified as nonsense if at least  \nclassifier replies in the ensemble predict the example is nonsense.
See Section~\ref{sec:appendix, baseline tuning} for details on tuning $t_r$ and \nclassifier.




\subsection{\ModelName: Automatically Generating Discriminative Replies} 
\label{sec:autoreply}
Generating hand-crafted replies requires significant human effort as it requires the human experts to manually design replies to cover various types of nonsense situations.
Additionally, many hand-crafted replies 
may not 
be sufficiently discriminative
(\eg \emph{``I don't understand''} can appear as a valid follow-up reply after many good messages, simply because the good messages are complex).



Our goal is therefore to automatically generate follow-up replies. This is challenging because we are searching within a combinatorially-large space for replies that are discriminative. In particular, we want to find replies that 
1) are \emph{not} likely after good messages (to avoid generic responses like ``sounds good'') and
2) are highly likely after \emph{multiple} bad messages (to avoid replies highly specific to one particular example, \eg \emph{``Germany is attacking Russia too, this was a really strange game.''}) 



Our proposed method, \ModelName, searches contrastively for replies that meet these criteria.
\ModelName is a pruned breadth-first search which constructs replies token-by-token recursively, only keeping partially-constructed replies that 1) maintain a probability margin between good and bad responses and 2) are high probability replies to a sufficiently large number of bad messages. 

As shown in Figure~\ref{fig:model}, the input to \ModelName is a dialogue model $\mathcal{L}$ which both generates and scores possible replies, a small set of annotated bad ($B$)  and good ($G$) examples with dialogue contexts to contrast the generated reply's probability against each other, and a set of search hyper-parameters ($p$, $K$, etc).
Each example consists of the current game state, the dialogue history, and the message to detect.
The output from \ModelName is a set of follow-up replies, such as \emph{``I can't do that''}.

\ModelName proceeds recursively, where each recursive step considers extensions to a prefix (which is initially empty when search begins; Figure~\ref{fig:model} shows an example for the prefix \emph{``I can't''}). See Algorithm~\ref{algorithm} in the Appendix for pseudocode.

\ModelName uses the following parameters: 
\begin{itemize}[leftmargin=*]
    \setlength\itemsep{-0.3em}
    \item $p$: a top-p parameter specifying the size of the token set for reply expansion. 

    \item $K$: the minimal number of bad examples a token has to appear in  for it to be expanded. $K$ controls the reply's specificity to individual examples. 
    
    \item $topn$: 
    To reduce the search space, similar to beam search, at each recursive call,  we sort the tokens by their frequency 
    across bad examples,  and expand only the $topn$ most-frequent tokens.
    


\end{itemize}
We now describe \ModelName's steps in detail.

\noindent\textbf{Obtain highly-likely token sets $V_i^{(p)}$.} Given a response prefix $r_0$,  we use the dialogue language model $\mathcal{L}$ to calculate the nucleus top-$p$ vocabulary set $V_i^{(p)}$ \cite{holtzman2019curious} of continuations to $r_0$ for each bad message example $B_i$. $V_i^{(p)}$ is the smallest set of tokens such that  
\begin{equation*}
    \sum_{v\in \sV_i^{(p)}} P_{\mathcal{L}}(v|B_i+r_0) \geq p
\end{equation*}
For instance, in Figure \ref{fig:model}, if we fix the reply prefix to be \emph{``I can't''}, for the first bad example $B_1$, the top-$p$ $\sV_1^{(p)}$ contain ``sup'' (the first subword token in ``support''), ``do'', ``move'' etc; for $B_i$, $\sV_i^{(p)}$ contain ``con'' (for convoy), ``trust'', etc. The advantage to obtaining a separate $V_i^{(p)}$ for each bad example is that we can attend to individual nonsense messages and produce diverse follow-up replies. For example, in our Diplomacy setting an agent could describe various kinds of invalid actions (\eg proposing to move immovable fleets, or armies, or support a partner that can't be supported), and therefore a single reply like \emph{``I can't move''} might not capture all the invalid order scenarios.
Obtaining $V_i^{(p)}$ for each bad example allows generating different replies like \emph{``I can't support''} and \emph{``I can't convoy''} 
to capture a broader range of nonsense.

\noindent\textbf{Choose common highly-likely tokens to expand.} 
After obtaining response continuation sets $V_i^{(p)}$ for each bad example $B_i$, the counts of tokens are aggregated across these sets.
The intuition for this aggregation is that we want to find replies that are general, rather than being highly-specific to particular bad examples.
We introduce the second parameter $K$ to achieve this. If a given token $v$ occurs in at least $K$ of the token sets $V$ (\ie it was highly likely for at least $K$ bad situations) then it will be expanded in the next step; otherwise 
$v$ will not be expanded, as it is too specific to particular examples. 
This parameter encourages follow-up replies to be generalizable across examples.

In Figure~\ref{fig:model}, the tokens ``do'', ``move'', ``sup'', ``con'', and ``trust'' appear in more than $K=15$ $V_i^{(p)}$ and therefore are candidate tokens to expand, while the tokens ``reach'', ``beat'', ``tell'' appear less than 15 times and are abandoned. 
We also record the set of bad examples each token appears in, $B_v=\{B_i: v\in\sV_i^{(p)}\}$, for the probability calculation in the following contrastive step.

\noindent\textbf{Contrast scores to find discriminative replies.} Our end-goal is to find \emph{discriminative} replies that can differentiate bad and good messages. 
But setting $p$ and $K$ cannot prevent non-discriminative generic replies such as ``sounds good'', which are highly-likely after any message. 
To address this, we contrast a partial reply's probability after bad messages with its probability after good messages to identify discriminative replies, and use this to prune the search.
As we did previously for bad examples, we also obtain a high-probability token set $V_{i, good}^{(p)}$ for each good example, and the set of good examples $v$ appears in, $G_v=\{G_i: v\in\sV_{i, good}^{(p)}\}$. 





We aggregate probabilities of the reply as a continuation across all good examples $G_v$, and contrast this aggregated probability with the aggregated probability across all bad examples $B_v$, as follows:
For $r=r_0+v$, like \emph{``I can't do''} in Figure~\ref{fig:model}, we obtain the set of its log probabilities conditioned on bad examples, $\log P_{B}(r)=\{\log P_{\mathcal{L}}(r|B_i)\mid B_{i}\in B_v \}$, and also the set of its log probabilities conditioned on good examples $\log P_{G}(r) = \{\log P_{\mathcal{L}}(r|G_i)\mid G_{i}\in G_v \}$.  
We use aggregation functions $f_b$ and $f_g$ (\eg mean, min, or max), to compute summary statistics of these log probabilities across the token sets for good and bad examples: $s_b(r) = f_b(\log P_{B}(r))$ and $s_g(r) = f_g(\log P_{G}(r))$. Given these summary statistics, we define a contrastive score $\Delta= s_b(r)  - s_g(r)$ 
and prune the search using this score and a threshold value $t_{\Delta}$.
For example, in Figure~\ref{fig:model} \emph{``I can't trust''} is pruned because its $\Delta\leq t_{\Delta}$. 
We give more details on these parameters in Section~\ref{sec:appendix, parameter tune}.




\noindent\textbf{Parameter tuning.} 
We tune the parameters of \ModelName by simulating the search on hand-crafted replies and looking for the set of parameters that prune the space to an affordable size while keeping the most hand-crafted replies. For more details, please refer to Section~\ref{sec:appendix, parameter tune}.  In our 
experiments,  we  use $T=6$ as the maximum length for the generated reply, $p=0.9$, $K=19$, $topn=15$, $\Delta_r=\text{mean}(\log P_{B}(r)) - \min(\log P_{G}(r)) > t_{\Delta}=3.63$. 


\noindent\textbf{Ensemble replies.}
To construct a nonsense classifier from each generated reply $r$, 
we  set $t_r = \max_i\{\log P_{\mathcal{L}}(r|G_i)\mid G_{i}\in G_v \}$, the maximum probability among good examples, as the probability threshold. If  for an example $x$, $P_\mathcal{L}(r|x)> t_r$,  we predict $x$ as nonsense; otherwise, it is good.

\begin{figure}
    \centering
    \includegraphics[scale=0.6]{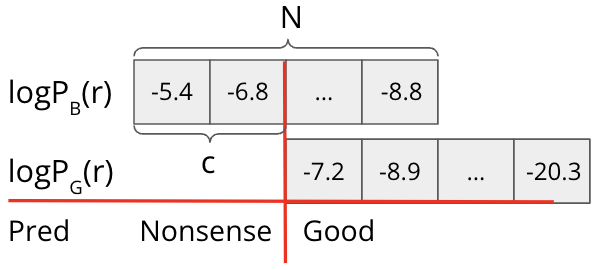}
    \caption{ Reply $r$'s log probabilities under bad and good examples. The $c$ bad examples with probability bigger than $\max(\log P_{G}(r))\}$ are predicted as nonsense. Thus, as a classifier, $r$'s precision=1, recall = $c/N$. Increasing $c$  leads to a smaller ensemble of replies with individual higher-recall above $c/N$.
    } 
    \label{fig:pred}
\end{figure}

As shown in Figure~\ref{fig:pred}, 
each $r$ is a classifier with precision=1 and recall $\geq c/N$ on the training set by construction, 
where $c$ is the number of bad examples whose probabilities are bigger than $t_r$, and $N$ is the total number of training bad examples. 
We can tune $c$ to get different subsets of the generated follow-up replies. 
A larger $c$  leads to a smaller ensemble of higher-recall replies whose individual recalls are $\geq c/N$. 
We show in Section~\ref{sec:classification result} that ensembling these  classifiers produces a high-precision, high-recall classifier.

\begin{table*}[!htbp]
\centering
\small
\begin{adjustbox}{width=0.95\textwidth}
\begin{tabular}{p{8em}p{22em}p{14em}}
\toprule
 \textbf{Nonsense type} & \textbf{Message} & \textbf{Human comment}  \\
 \midrule
 
 Wrong justification & \germany $\rightarrow$ \england: i'm interested, but don't tell france that. i'll move my fleet to hel, so that i can take belgium and then start moving armies east & 
Moving to hel doesn't help with taking Belgium.\\

\midrule


Invalid order proposal (for listener) & \russia $\rightarrow$ \germany: Are you moving in from Norway or the Barents Sea?
& Germany doesn't have a unit in Barents.\\
\midrule

Invalid order proposal (for self) & \england $\rightarrow$ \germany: so. would you like support in to sweden from norway?
& England can't support this move.\\

\midrule



Contradiction (with game state) & \russia $\rightarrow$  \italy: You should have taken Marseilles when you had the chance & Italy has Marseilles. \\

\midrule 

 




 General nonsense & \austria $\rightarrow$ \italy: Sorry, the webpage keeps sending duplicate messages. & Austria did not send a duplicate message. \\ 


\bottomrule

\end{tabular}
\end{adjustbox}
\caption{ Examples of annotated nonsensical message in Diplomacy: while these messages contain reasonable surface forms, expert humans with access to the game state can tell that they are nonsensical.}
\label{tab:example messages}
\end{table*}

\begin{table}[htb!]
\centering
\small
\begin{adjustbox}{width=0.9\columnwidth}
\begin{tabular}{lrrr}
\toprule
 \textbf{Label} & \textbf{Train} & \textbf{Validation} & \textbf{Test} \\
 \midrule
Good (88\%)& 4,149 & 518&518\\
Nonsense (12\%) & 561 & 69 & 70\\
\midrule
Total & 4,710 & 587 & 588\\
\midrule\midrule
\textbf{Avg \# Msg in Context} & \textbf{Train} & \textbf{Validation} & \textbf{Test}\\
\midrule
\DataName& 140.2&148.0&139.8\\

ConvAI2 \cite{dinan2019second}& 7.5 & 7.8 & -\\

LIGHT \cite{urbanek2019learning}& 9.8 & 9.8 & 9.8\\

\bottomrule
\end{tabular}
\end{adjustbox}
\caption{ \textbf{\DataName} dataset statistics. 
The classes are highly imbalanced and the context is long. 
}
\label{tab:dataset stats}
\end{table}

\section{Diplomacy and Data Collection}
\label{sec:diplomacy_data}

We evaluate our method on Diplomacy, a seven-player board game where each player controls the units (fleets and armies) of an European power starting in the year 1901, with the goal  to win as many \emph{supply centres} as possible. In \cref{appendix:rules}, we briefly describe the game rules. One game consists of many years, and each year is divided into phases between which, the players are permitted to communicate with each other, and thus the conversations are very long:  
on average, each training example's dialogue history contains 140+ messages richly grounded on the game state (see Table~\ref{tab:example messages} for examples).  As a comparison, the average number of messages in context for ConvAI2 \cite{dinan2019second} (a widely-used dialogue dataset) and  LIGHT \cite{urbanek2019learning} (a dialogue dataset on a fantasy text game) ranges from 7 to 9 messages.




 Now we introduce \DataName, the dataset we collected for the nonsense detection task. We had experienced Diplomacy players follow the taxonomy in Table~\ref{tab:example messages} to  annotate nonsensical messages produced by the BART dialogue agents (Section~\ref{sec:experiments})  in self-play. 
We chose self-play games because they are more likely to contain nonsense than human-human games. 
The annotation process is labor intensive as the games are long and it takes much time  even for human experts to understand the changing game state and the complex conversation history. Table~\ref{tab:example messages} shows example messages which all look  reasonable on the surface but are actually nonsensical. For instance, in the first example, Germany says to England, \emph{``i'll move my fleet to hel [Helgoland Bight], so that i can take belgium''},  but under the current game state, moving to Helgoland Bight doesn't help take Belgium; detecting this requires domain knowledge which novice players typically lack. There are also many nonsense types, such as wrong justification of previous movements, invalid order proposal for other players or themselves,  contradiction, etc, and each type requires its own replies.
Table~\ref{tab:dataset stats} shows the \DataName statistics, with highly imbalanced  classes  (only $12\%$ messages are nonsense), making the nonsense detection more challenging. 

In summary, \DataName is a challenging dataset, in which detecting nonsensical message can involve both reasoning over long dialogue contexts and grounding in a rich environment. See \cref{appendix:dataset} for more details on the dataset.

\begin{table*}[!htbp]
\centering
\begin{adjustbox}{width=0.95\textwidth}
\begin{tabular}{lrrrrrrrrr}
\toprule
 && 
\multicolumn{4}{c}{\textbf{Validation}}  & 
\multicolumn{4}{c}{\textbf{Test}} 
\\ 
\cmidrule(lr){3-6} 
\cmidrule(lr){7-10} 
 \textbf{Model} & \textbf{num} &  
 \textbf{Auc}& \textbf{Prec} & \textbf{Recall}&\textbf{F1}&  
 \textbf{Auc}&  \textbf{Prec} & \textbf{Recall} & \textbf{F1}  
 \\
\midrule


Hand-crafted replies&14&
59.81&22.52&36.23&27.78&
58.73&20.31&37.14&26.26\\ 

\midrule

$\mathcal{L}$-generated replies&8834&
58.94&24.42&30.44&27.10&
59.05&24.18&31.43&27.33\\ 

\midrule

\ModelName (num=14) 
&14&
48.23&10.66&30.44&15.79&
63.36&19.90&\textbf{58.57\rlap{$^*$}}&29.71\\ 






\ModelName (num=2805) 
&2805&
\textbf{63.58}&\textbf{27.36}&42.03&\textbf{33.14}&
\textbf{67.12\rlap{$^*$}}&\textbf{28.80\rlap{$^*$}}&51.43\rlap{$^*$}&\textbf{36.92\rlap{$^*$}}\\ 

\ModelName (num=2805, picked) 
&82&
60.89&17.00&\textbf{62.32\rlap{$^*$}}&26.71&
57.63&16.28&50.00\rlap{$^*$}&24.56\\ 


\midrule
\midrule
Supervised Learning &-& 
70.06\rlap{$^*$} & 24.47 & 68.12\rlap{$^*$} & 36.02\rlap{$^*$} &
71.85\rlap{$^*$}  &25.24 &  72.86\rlap{$^*$} &37.50\rlap{$^*$} \\%

\bottomrule
\end{tabular}
\end{adjustbox}
\caption{ Main  classification results.  ``\ModelName (num=$x$)'' gives results for an ensemble of $x$ replies generated by \ModelName. 
 ``\ModelName (num=2805)'' achieves significantly better classification results than all the baselines. ``\ModelName (num=14)'' contains only 14 replies, and is still better than the baselines, suggesting the generated replies are of high-quality.
  ``\ModelName (num=2805, picked)'' is a subset of ``\ModelName (num=2805)'' selected using human-defined keywords like ``can't'', suggesting that integrating human knowledge doesn't work. 
 * indicates results that are statistically significant in comparison to hand-crafted using a paired sample bootstrap test.
}
\label{tab:main}
\end{table*}

\section{Experiments}
\label{sec:experiments}

In our experiments, we fine-tune a case-insensitive BART \cite{lewis2019bart} $\mathcal{L}$  on human-human Diplomacy dialogues from WebDiplomacy\footnote{https://webdiplomacy.net/}    and use it to both score and generate the follow-up replies. 
For more experimental details, please refer to Section~\ref{sec:appendix, baseline tuning}. 
We compare \ModelName's classification results against various baselines for nonsense detection and analyze the generated replies qualitatively. We also compare it against a supervised model that requires training a large-scale classifier. Now we describe the models.


\noindent \textbf{Hand-crafted}, is a baseline classifier built with the hand-crafted replies, as described in Section~\ref{sec:baseline}. 

\noindent \textbf{$\mathcal{L}$-generated replies},  where we use the language model $\mathcal{L}$ directly to generate 20 replies for each bad example and ensemble them  for a simple baseline.  $\mathcal{L}$-generated replies  are often specific to one single bad example and ignore the nonsense-related aspects, e.g., \emph{``can you please support albania to greece?''}, \emph{``russia lied to me, as he is working with turkey. we can win if we work together.''}, etc. 
For the threshold, for a fair comparison with \ModelName, we also use the maximum log probability over good training examples $t_r = \max_i\{\log P_{\mathcal{L}}(r|G_i)\mid G_{i}\in G_v \}$.

\begin{table}[!htbp]
\centering
\begin{adjustbox}{width=0.95\columnwidth}
\begin{tabular}{llcccc}
\toprule
\multirow{2}{45mm}{\textbf{Model on ``Invalid Order'' (79 training examples)}}
&&  
\multicolumn{4}{c}{\textbf{Test (518/14)}} \\ 
\cmidrule(lr){3-6} 
 &\textbf{num}&  
 \textbf{Auc}&  \textbf{Prec} & \textbf{Recall} & \textbf{F1}  \\

\toprule
Hand-crafted&5&
57.92&9.38&21.43&13.04\\ 
\midrule
\ModelName (num=1609) 
&1609&
60.23&37.50&21.43&27.27\\ 

\midrule
\midrule

Supervised Learning  &-& 
74.42
& 10.11 
& 64.29
& 17.48\\%

\bottomrule
\end{tabular}
\end{adjustbox}
\caption{Classification results on the ``invalid order'' subset with 79 examples. \ModelName performs better than the hand-crafted baseline, and the supervised model.}
\label{tab:invalid order in main}
\end{table}

\noindent\textbf{\ModelName (num=$x$)}. 
 \ModelName generates many replies and as mentioned in  reply ensembling in Section~\ref{sec:autoreply}, we can tune $c$ to obtain different subsets of replies. ``\ModelName (num=$x$)'' is a subset of \ModelName-generated replies with $x$ replies  whose individual recalls are above a threshold. 
 

\noindent\textbf{Supervised classifier.} As a reference, we extensively tuned the hyperparameters to obtain a large BART-based supervised classifier on ($B$, $G$). It conditions on the same information (game state, message history, etc) as the dialogue model $\mathcal{L}$. 

 
\textit{Data used in different phases.} Using the probability of replies for nonsense detection requires  tuning certain parameters (probability threshold $t_r$, \nclassifier for the ensemble, etc) and we tune them on different data. In summary, we generate the replies and tune the probability thresholds on the train set to construct the replies as individual classifiers, 
and then tune the ensemble parameter \nclassifier on the validation set. Table~\ref{table: data explanin} summarizes the  data used in different phases for different models. 

\begin{table}[!ht]
    \centering
    \begin{adjustbox}{width=0.95\columnwidth}
    \begin{tabular}{l|l|l|l}
    \toprule
        \textbf{Model} & \textbf{Reply Generation} & \textbf{Prob Threshold} & \textbf{n\_classifier}  \\ \midrule
        Hand-crafted reply & Manual & all Train & Validation \\ \midrule
        $\mathcal{L}$-generated reply & 561 bad in Train & all Train & Validation \\ \midrule
        \ModelName & \multicolumn{2}{c|}{561 bad + 561 good in train} & Validation \\ 
        \midrule
        Large supervised & \multicolumn{3}{c}{561 bad + 561 good in train + Validation}  \\ 
        \bottomrule
    \end{tabular}
    \end{adjustbox}
    \caption{Data used in different phases.}
    \label{table: data explanin}
\end{table}

\subsection{Classification Result}
\label{sec:classification result}
Table~\ref{tab:main} shows the main classification results. 
The hand-crafted baseline achieves a test F1 of $26.26$, indicating that the hand-crafted  replies are able to detect nonsense even in a complex application like Diplomacy (as a comparison, majority vote results in a test F1 of $0$).  
``\ModelName (num=2805)'' achieves a test F1 of $36.92$ with 2805 follow-up replies, significantly higher than the hand-crafted baseline ($26.26$) and $\mathcal{L}$-generated baseline ($27.33$), and comparably to the model trained with supervised learning ($37.50$). This shows that \ModelName can automatically generate large numbers of replies to improve nonsense detection. 

The best hand-crafted reply set contains 14 replies. To control the effect of reply amounts, we also get a smaller subset with only 14 replies from \ModelName (``\ModelName (num=14)''), by increasing $c$ in Figure~\ref{fig:pred} as mentioned earlier.  
It achieves a better test F1 of $29.71$ with the same number of replies.
This shows that when we control the  reply amount, \ModelName can produce higher-quality discriminative replies. 
Compared to \ModelName, although ``$\mathcal{L}$-generated'' also generates a large number ($8834$) of replies,
it achieves a much lower test F1 ($27.33$), because most $\mathcal{L}$-generated replies are too specific to the bad examples in the train set. 
This suggests that  the reply quality matters more than the quantity.

Next, we explore if manual curation can select an even higher quality set of replies from those found by \ModelName.
We define a set of keywords (e.g., \emph{``can't''}, \emph{``makes no sense''}) to filter a subset of the replies from ``\ModelName (num=2805)'', denoted as ``\ModelName (num=2805, picked)''. 
This leads to a much smaller set with only $82$ replies and a lower test F1 of $24.56$, comparable to the hand-crafted baseline, further confirming that manual reply design is challenging. 

Now we compare \ModelName with large supervised models. We note that the supervised models are extensively fine-tuned on the task specifically. 
Table~\ref{tab:main} shows that on the whole dataset, \ModelName performs on par with the large supervised model, with similar F1 ($36.92$ vs $37.50$, the difference is \emph{not} significant) and slightly better precision ($28.80$ vs $25.24$). We also perform classification on the ``invalid order'' nonsense subset (Table~\ref{tab:invalid order in main}). This is an important subcategory because proposing invalid orders shows the agent is not familiar with the game, and would most likely hurt the dialogue agent's credibility in games with human Diplomacy experts.
\ModelName achieves a better test F1 ($27.27$) than the supervised model ($17.48$), demonstrating that \ModelName may work better than supervised models in low-resource settings. These promising results show that \ModelName at least matches a large fine-tuned supervised model. 

For more related low-resource results, please see Section~\ref{sec: more experiments}. We also show  single replies without ensemble also achieve promising results in Section~\ref{sec:single-reply}. We plan to develop  better ensemble methods to further improve the ensemble performance.

\subsection{Qualitative Analysis}

We also analyze the generated replies qualitatively. Table~\ref{tab:example generated replies} shows  \ModelName-generated replies. 
We manually clustered them for intepretability. 

We observe that the generated replies are diverse (e.g., \emph{``you are hitting refresh''} and \emph{``triple posts are strange''} for repetition), and cover different types of nonsense. 
These examples also demonstrate \ModelName's potential use as a paraphrase generation tool: given the same prefix, it could find semantically-similar tokens, e.g., \emph{``i don't understand your messages''}, and \emph{``i don't understand your point''}; \emph{``you have no fleets''}, and \emph{``you have no troops''}. Some generated messages are not decoded to the end intentionally to reduce the computational cost. Also, the sentences don't have to be complete to be discriminative, e.g., \emph{``i think you meant a different''} could be followed by ``country'' or ``player''. If the incomplete sentences are already discriminative enough, we can still utilize their probabilities to detect nonsense.

\begin{table}[!htbp]
\centering
\small
\begin{adjustbox}{width=0.95\columnwidth}
\begin{tabular}{lm{50mm}}
\toprule
 \textbf{Label} & \textbf{Message}  \\
\midrule
\multirow{3}{*}{to wrong country} & i think you meant to send
\\
&  what? why did you send\\
& i think you meant for someone\\
& i think you meant a different\\

\midrule

\multirow{4}{*}{general nonsense} &  no, that makes no sense
\\
&  i don't understand your messages\\
& i don't understand your point\\
& what are you talking about???\\

\midrule

\multirow{5}{*}{self invalid order} &  i think you can't move
\\
& how? you have no fleets\\
&how? you have no troops\\
& you can't do that because\\

\midrule

\multirow{5}{*}{other invalid order} &  no, i can't move
\\
&  no, i can't sup\\
&  yeah but it doesn't work\\
& i am sorry i can not\\

\midrule

\multirow{5}{*}{repetition} &  you have triple messages
\\
& you've said that before\\
& you are hitting refresh.\\
&triple posts are strange\\

\bottomrule

\end{tabular}
\end{adjustbox}
\caption{ \ModelName-generated reply examples. The replies are diverse and cover different nonsense types. }
\label{tab:example generated replies}
\end{table}

\begin{table*}[!htbp]
\centering
\begin{adjustbox}{width=0.95\textwidth}

\begin{tabular}{lrrrrrrrrrrrrrr}
\toprule
& & & & & & 
\multicolumn{4}{c}{\textbf{Validation}}  &
\multicolumn{4}{c}{\textbf{Test}} 
\\ 
\cmidrule(lr){7-10} 
\cmidrule(lr){11-14} 
\textbf{Model}&\textbf{ng}&\textbf{p}& \textbf{K}&\textbf{topn}&\ \textbf{num}&
  \textbf{Auc}& \textbf{Prec} & \textbf{Recall}&\textbf{F1}&  
  \textbf{Auc}&  \textbf{Prec} & \textbf{Recall} & \textbf{F1}  
  \\

\toprule







\ModelName (num=6700)& 561& 0.9 & 19 & 15 & 6700&
61.84&27.78&36.23&31.45&
66.43&30.84&47.14&37.29\\ 
\midrule
\ModelName (ng=50)
&\textbf{50} & 0.9 & 19 & 15 & 6743&

63.53&32.10&37.68&34.67&
56.99&20.19&30.00&24.14\\ 

\ModelName (p=0.8) 
& 561& \textbf{0.8} & 19 & 15  & 4282&
62.68&17.69\rlap{$^*$}&66.67&27.96&
62.12&17.13\rlap{$^*$}&70.00&27.53\rlap{$^*$}\\ 

\ModelName (K=7)
&561 & 0.9 & \textbf{7} & 15  &11138&

59.19&23.91&31.88&27.33&
56.97\rlap{$^*$}&22.79\rlap{$^*$}&25.71\rlap{$^*$}&24.16\rlap{$^*$}\\

\ModelName (topn=10)
& 561 & 0.9 & 19 & \textbf{10}  &6312&

60.29&29.17&30.44&29.79&
61.45\rlap{$^*$}&28.92&34.29\rlap{$^*$}&31.37\rlap{$^*$}\\ 



\bottomrule
\end{tabular}
\end{adjustbox}
\caption{Classification result of \ModelName with different parameters. \textbf{ng}: number of  good examples. We perform a paired sample bootstrap test against ``\ModelName (num=6700)'' and * indicates significantly lower (worse) results. Lowering $p$, $K$ and $ng$ negatively impacts the results as it leads to less-diverse or too-specific replies. 
}
\label{tab:ablation}
\end{table*}

Our main goal is to utilize the dialogue model’s own introspective ability to detect its own mistake, without building another model or adding more parameters. Prefix-tuning with non-human-readable prompts is worth investigating as a parallel future direction, but it still adds extra parameters to the dialogue model. Our goal is to explore the potential of the dialogue model itself to introspect.


\subsection{Parameter Analysis}
Table~\ref{tab:ablation} shows \ModelName's sensitivity to its parameters.  
For fair comparison, we keep the number of replies on a  similar scale across conditions. 

We first investigate the importance of the number of good situations used to contrast against. The second row in Table~\ref{tab:ablation} shows that classification performance drops substantially when the number of good examples is reduced $50$, as the algorithm finds more spurious correlations between replies and nonsense annotations.
For example, \ModelName  generates \emph{``thanks turkey''} as a discriminative reply because in the bad examples, many messages were sent to Turkey  
but in the 50 good examples, none of the messages were sent to Turkey, so \ModelName considers ``Turkey'' as a discriminative token that only appears in bad examples. 

Next, we decrease $p$ from $0.9$ to $0.8$ (which reduces the number of tokens explored in each $V_i^{(p)}$), causing generated replies to be less diverse. We find that this in turn lowers classification performance. 
Decreasing K from $19$ to $7$ (which allows finding replies which are likely after fewer bad examples) causes generated replies to be highly specific to particular examples, \eg \emph{``no. France can support''}, also lowering performance.  
Decreasing $topn$ from $15$ to $10$ reduces the search space, but also leads to more generic responses and a lower test F1 of $31.37$. However, the impact of changing $topn$ only is smaller than changing $p$ or $K$.

\section{Related work}
Detecting nonsensical dialogue 
is a well-known challenge \cite{li2019don, shi2020refine}. 
Previous research formulated nonsense detection as a supervised learning problem \cite{welleck2018dialogue,nie2020like}. 
But as collecting such datasets can be costly and difficult to scale, recent work proposes to evaluate generated text with prompt-based learning. 
\citet{yuan2021bartscore} proposed BARTScore and showed that the probability of hand-crafted prompts can be used for text quality evaluation. 
\citet{mehri2020unsupervised} used DialoGPT \cite{zhang2019dialogpt} to get the probability of hand-crafted follow-up replies 
to evaluate the generated dialogues. 
We also use follow-up replies to detect bad messages, but instead of hand-crafted replies, \ModelName automatically generates many follow-up replies  to cover different nonsense types.

Our work is one type of ``prompt-based learning'', which utilizes pretrained language models and text prompts for downstream NLP tasks \cite{shin2020autoprompt, liu2021pre}.  
While prompts can be manually designed \cite{petroni-etal-2019-language,brown2020language},
recent work has proposed prompt generation \cite{jiang2020can,li-liang-2021-prefix}
to automatically obtain high-quality prompts at scale.
For instance, \citet{wallace-etal-2019-universal} proposed a gradient-guided method to search for tokens that would trigger certain targets, which requires many training examples.  
\citet{gao2020making} used limited annotated examples to predict tokens at specified positions in the template. 
Different from previous work, our goal is to generate free-form follow-up replies that are \emph{discriminative}: likely after nonsensical messages but unlikely after good messages. 

Diplomacy has been a long-standing AI benchmark game due to its large action space and complex communication between players. Most relevant previous studies focused on the ``no-press'' variant of the game, 
where communication is not permitted. \citet{paquette2019no} presented the first neural-network-based policy model for no-press Diplomacy. 
\citet{bakhtin2021no} trained a no-press Diplomacy agent without human data and achieved super-human performance in a popular two-player Diplomacy version. 
\citet{jacob2021modeling} built agents for several games including no-press Diplomacy that are simultaneously strong and human-like. 

Research on full-press Diplomacy, the variant that permits communication, has been limited. \citet{niculae2015linguistic} explored deception dynamics in Diplomacy, while \citet{peskov2020takes} studied  the use of lies in Diplomacy and collected a dataset with truth and lie message annotations. In this paper, we focus on Diplomacy as a complex testbed for nonsensical dialogue message detection.  

\section{Conclusions and Discussion}



In this paper, we 
propose to detect nonsensical messages using the probability of the dialogue model's own follow-up replies  like \emph{``I don't understand''}, without building an extra classifier. We evaluate this reply-based approach on Diplomacy, a complex board game with rich verbal communication dynamics. We first show that hand-crafted replies are effective for nonsense detection. To reduce the labor of reply engineering, we develop \ModelName, a search algorithm to automatically generate discriminative replies. Experiments show that \ModelName can generate many high-quality discriminative replies and  achieves significantly better performance than the hand-crafted baselines, and performs on par with large supervised models.   

One thing to note is that our reply-based approach and \ModelName are not limited to Diplomacy, and can be applied to various other dialogue problems, such as improving dialogue safety by detecting offensive language (with replies like \emph{``that's not nice to say''}) 
or detecting 
factual contradictions. We acknowledge that \ModelName still have much potential for improvements. For
for limitations, ethical considerations and future work, please refer to Section~\ref{sec: ethical} and~\ref{sec:future work}.



\newpage

\bibliography{anthology,custom}
\bibliographystyle{acl_natbib}
\appendix

\newpage
\section{Appendix}
\label{sec:appendix}

\subsection{Limitations and Ethical Considerations}
\label{sec: ethical}
We acknowledge that \ModelName has a few limitations that may make it less effective in certain settings.  As mentioned earlier, \ModelName still requires annotated examples and its performance relies on the quantity and the quality of the contrastive good examples. If only limited good examples are available, then \ModelName may not perform as well. 

Additionally, computing the probability for generated replies during the search is computationally expensive: on average, generating \ModelName requires 200 32GB V100 GPUs for roughly 24 hours. Moreover, classification with \ModelName requires calculating the probability of each reply: for the whole ensemble, it takes on average 200 32GB V100 GPUs roughly 1 hour on the train and validation sets.  More work needs to be done to develop methods for making this more efficient (please see future work in Section~\ref{sec:future work}). 

Previous work has noted that various harms may result from interacting with dialogue agents, e.g., \citet{dinansafety2021}. We note that the creation of \DataName did not involve human interactions with dialogue agents: rather, dialogues were generated through self-play (i.e., dialogue agents interacting with each other), and subsequently, these generated dialogues were annotated by humans for nonsense. The annotators used in the creation of this dataset were members of the authors' lab with experience playing the game Diplomacy.

Diplomacy is a board game, and as such, nonsense detection in this specific domain may have limited real-world utility. However, the methods used in this paper are not specific to Diplomacy and are therefore generalizable to the detection of ``bad'' messages in other settings involving dialogue agents. We can imagine that such techniques might help with applications such as detecting misinformation from dialogue agents \citep{weidingersocialrisks2021}, or even, language models' self-diagnosis of toxic generations \citep{selfdiagnosis2021schick}.  Applications of these techniques will need to ensure fairness of classification predictions.

Finally, as noted, creating and using \ModelName incurs a high computation, and therefore, environmental cost \citep{strubell2019environment}.  More work needs to be done to improve the efficiency of techniques used in this paper before they can be widely applied in other settings.

\subsection{Future Work}
\label{sec:future work}

\noindent\textbf{Better ensemble methods for both better performance and faster inference. }
As show in the following Section~\ref{sec:single-reply}, \ModelName can generate many effective single replies (e.g. a single reply without ensemble can achieve a test F1 of $29.17$), and currently we use a simply voting-based ensembling scheme: simply if more than N replies in the ensemble decide the example to be nonsense, then the final result will be nonsense. But given the complex dialogues, these replies can often conflict with each other.  If more sophisticated ensemble methods using fewer replies are developed (e.g., choose the top 50 performing replies and ensemble them smartly for different examples to avoid conflicts), AutoReply has a big potential to improve the ensemble performance further with less computation cost during inference.

Also, now we tune the ensemble hyparameters according to the validation set. But given the complex game situation, the validation set and test set are relatively small, and the validation is not necessarily reflective of the test set (Table~\ref{tab:single} in Section~\ref{sec:single-reply}).  We plan to study how to more effectively utilize the validation set given limited data in the future. 

\noindent\textbf{Speed up \ModelName. } To speed up the \ModelName in the future, we plan to store intermediate probability, more effectively utilize the contrastive good example set, and prune the search space more with different sets of parameters. One advantage of \ModelName is that it provides multiple hyperparameters so that the users can prune the spaces differently according to the computation resources they can afford. In our experiments, we chose a setting according to our available computation resources.  
In practice, we find using smaller maximum length $T$ and $topn$ prunes the space the most, and we plan to perform more extensive parameter tuning on them to see the effects. 

\noindent\textbf{Parallel research directions. } Our goal is to utilize the dialogue model to introspect and find its own mistakes without extra models or extra parameters. In the future, we also plan to explore parallel research directions in the space of dialogue nonsense detection, such as prefix tuning which replies on non-human-readable prompts, and paraphrase tools which can potentially enrich the hand-crafted replies.

\subsection{How to tune the parameters}
\label{sec:appendix, parameter tune}

To prune the space, we set a maximum length $T$ for the generated responses, and sort the tokens by its occurrence in bad examples and expand only the $topn$ most-frequent tokens. 
Also, different contrastive score $\Delta_r$ and $t_{\Delta}$  prune the spaces in different ways. 
To tune the parameters, we simulate the search with the hand-crafted replies, and select $T$, $p$, $K$, $topn$, $\Delta_r$, $t_{\Delta}$, $f_b$ and $f_g$ 
that  prunes the space to an affordable size while keeping the most hand-crafted replies.  $f_b$ and $f_g$ could be $\min$, $\max$, mean, the n-th biggest value, the mean over the topn biggest values, etc. 
 
In our best experiments,  we start the prune when $len(r) >= t_{\text{prune}}=3$, 
and use $T=6$, $p=0.9$, $K=19$, $topn=15$, $\Delta_r=mean(\{\log p_{b_{i}}(r)|b_{i}\in B_v \}) - \min(\{\log p_{g_{i}}(r)|g_{i}\in G_v \}) > t_{\Delta}=3.63$ 
and seven good hand-crafted replies can be kept with this set of parameters.

\begin{table*}[!htbp]
\centering
\begin{adjustbox}{width=0.95\textwidth}
\begin{tabular}{m{12mm}|m{20mm}|m{50mm}|ccc|ccc}
\toprule
\textbf{Data}&\textbf{Model} &\textbf{Best Reply (ordered by valid result)}& \multicolumn{3}{c|}{\textbf{Valid}}&\multicolumn{3}{c}{\textbf{Test}} \\ 
\cmidrule(lr){4-9} 
&&&  \textbf{Prec} & \textbf{Recall} & \textbf{F1}&\textbf{Prec} & \textbf{Recall} & \textbf{F1}  \\
\midrule


\multirow{6}{10mm}{Full data} & \multirow{3}{30mm}{Hand-crafted} &  that's stupid  & 
17.14& 60.87& 26.75& 15.25 & 51.43 & 23.53\\ 
&&what are you talking about& 
 16.93& 62.32& 26.63&
15.58& 61.43& 24.86\\
&&you just said that& 
15.00& 86.96& 25.59&
14.08& 82.86& 24.07\\
\cmidrule(lr){2-9} 




&\multirow{3}{30mm}{\ModelName} & that is an excellent idea actually & 
17.91&34.78&23.65&

20.59& 50.00& 29.17\\
&&yes. france can take&
28.21 & 15.94& 20.37&

18.92& 10.00& 13.08\\
&&that is true! i will & 
25.00& 15.94& 19.47&
32.65 & 22.86 & 26.89\\
\cmidrule(lr){2-9} 


&Supervised&- & 24.47 & 68.12 & 36.02 & 25.24 & 72.86 & 37.50\\

\midrule

\multirow{6}{10mm}{``Invalid order'' subset}&\multirow{3}{30mm}{Hand-crafted}&i can't reach &

25.00& 11.11& 15.39&

33.33 & 21.43 &  26.09\\
&&you don't have any units there &
25.00& 11.11& 15.39&
50.00 & 7.14 & 12.50\\
&&you can't reach & 
10.53&22.22 & 14.29&
10.00 & 14.29 & 11.77\\
\cmidrule(lr){2-9} 


&\multirow{3}{30mm}{\ModelName}&how about i convoy & 
66.67& 22.22& 33.33&
40.00 & 14.29 & 21.05\\
&&how about if i con &
66.67& 22.22& 33.33&
40.00 & 14.29 & 21.05\\

&&how about i con & 
66.67& 22.22& 33.33&
40.00 & 14.29 & 21.05\\
\cmidrule(lr){2-9} 
&Supervised&- & 4.94 & 44.44 & 8.89 & 10.11 & 64.29 & 17.48\\

\bottomrule
\end{tabular}
\end{adjustbox}
\caption{Single-reply classification result.}
\label{tab:single}
\end{table*}

\subsection{Additional Experimental and Dataset Details}
\label{sec:appendix, baseline tuning}
\label{appendix:dataset}
\par{\textbf{Dataset}} \DataName was collected based on 13 self-play games produced by dialogue chatbots (not the subject of this work) built for the Diplomacy domain. The chatbots are based on a BART model fine-tuned on human-human Diplomacy dialogues from WebDiplomacy\footnote{https://webdiplomacy.net/}, by conditioning on game state, conversation history and other metadata. The agents playing the 7 powers were trained on only the dialogue histories that the specific power participates in, so as to prevent any data leakage between powers.

\noindent\textbf{Hand-crafted reply ensemble details.} We use the training set to tune the probability threshold for each hand-crafted reply: we prepare a set of log probability threshold candidates ranging from $-5$ to $-30$ with the spacing to be $0.5$ (i.e., $-5.5$, $-6$, ..., $-29.5$, $-30$), and  calculate the train F1 scores for each threshold candidate and choose the one with the best train F1 as the threshold $t_r$ for each hand-crafted reply $r$.


Each hand-crafted reply in the ensemble is a weak classifier and they can conflict with each other, so different subsets of them will lead to different results. So we sort the hand-crafted replies based on their train F1 score from high to low, and then try different subsets of them (top1, top2, ..., until all of them) and choose the subset with the best validation F1 and present its results in the Experiment section. 

We tune the ensemble parameter \nclassifier on the validation set. Basically, the parameters of individual classifiers (e.g., $t_r$) are tuned using the train set because the train set contains more examples than validation set, so the individual parameters can represent more data. Any ensemble parameters \nclassifier are tuned using the valid set because both validation and test sets are unseen, and tuning the ensemble parameters on validation is more indicative of the test performance than tuning on train.



\subsection{Single-reply Result}
\label{sec:single-reply}
In this section, we analyze the result using one single-reply instead of an ensemble of replies, which would save much computation cost at inference time. We pick the top-3 replies that perform the best on the valid set, and show their test performance in 
Table~\ref{tab:single}. The replies are listed in the order of their validation performance. The conclusion is that \ModelName can find high-quality discriminative replies which achieve reasonable performance by themselves (without ensembling them). Future research on how to better ensemble them and how to better utilize the validation set could improve the \ModelName ensemble performance further. 

For the full data performance, 
The best hand-crafted reply is \emph{``that's stupid''} with a test F1 of $23.53$. 
The best \ModelName reply is \emph{``that is an excellent idea actually''} and its test F1 is $29.17$,  higher than the best hand-crafted reply, which indicates that \ModelName can generate high-quality discriminative replies. 
Although this reply might not be as ``discriminative'' as \emph{``that's stupid''} from humans' perspective, it has a better 
test classification result, which shows the \ModelName is doing its job in finding truly ``discriminative'' replies according to the data, 
and suggests again that human knowledge might introduce biased stereotypes.

For the ``invalid order'' low-resource subset, the best hand-crafted reply is \emph{``i can't reach''} with a test F1 of $26.09$, and the best \ModelName-generated reply is \emph{``how about i convoy''} with a test F1 of 21.05, better than the supervised model whose F1 is 17.48.  

We note that the dataset size is small relative to the complex situations in Diplomacy, so the validation performance is not very representative of the test set. For example, ``yes. france can take'' achieves a valid F1 of 20.37, but its test F1 is only 13.08. Future research should also focus on how to utilize the validation set more effectively given the dataset size.

\begin{table*}[!htbp]
\centering
\begin{adjustbox}{width=0.95\textwidth}
\begin{tabular}{llcccc|cccc}
\toprule
 &&
\multicolumn{4}{c}{\textbf{Validation (518/9)}}  &
\multicolumn{4}{c}{\textbf{Test (518/14)}} \\ 
\cmidrule(lr){3-6} 
\cmidrule(lr){7-10} 
\textbf{Model} &\textbf{num}&  
 \textbf{Auc}& \textbf{Prec} & \textbf{Recall}&\textbf{F1}&
 \textbf{Auc}&  \textbf{Prec} & \textbf{Recall} & \textbf{F1}  \\

\toprule
Hand-crafted&5&
69.62&12.90&44.44&20.00&
57.92&9.38&21.43&13.04\\ 
\midrule
\ModelName (num=4, lumped order) 
&4&
48.94&0.00&0.00&0.00&
59.85&25.00&21.43&23.08\\ 
\ModelName (num=1609, lumped order) 
&1609&
55.46&50.00&11.11&18.18&
60.23&37.50&21.43&27.27\\ 


\midrule

\ModelName (num=23, lumped order) 
&23&
54.98&14.29&11.11&12.5&
56.47&22.22&14.29&17.39\\


\ModelName, fine-grained self invalid + other invalid
&19&
54.78&11.11&11.11&11.11&
56.37&20.00&14.29&16.67\\ 

\midrule
\midrule

Supervised Learning  &-& 
64.79  & 4.94  & 44.44\rlap{$^*$} & 8.89 &
74.42\rlap{$^*$} & 10.11 & 64.29\rlap{$^*$} & 17.48\\%

\bottomrule
\end{tabular}
\end{adjustbox}
\caption{Classification results on the subset of ``invalid order'' nonsense. We only have 79 annotated training examples so this is under low-resource setting. We also have more fine-grained annotations of ``self invalid order'' and ``other invalid order''. We apply \ModelName on each fine-grained category to generate replies and combine them (``\ModelName, fine-grained self invalid + other invalid'') to compare against the case where we lump the two categories together (``\ModelName (num=23, lumped order)''). \ModelName is still better than hand-crafted baselines. Lumping the categories together leads to more training examples, and thus more better replies and better classification results.  We perform t-test against ``Hand-crafted''. 
}
\label{tab:invalid order}
\end{table*}

These results show that a single-reply, which means doing one inference without much computation cost, can also lead to reasonable performance. 
\wyshi{But we also note that each reply is a weak classifier and sometimes they can conflict with each other. In our approach, we use a simple voting-base ensemble mechanism to ensemble these replies and make the final prediction. Utilizing more sophisticated ensemble approach in the future could potentially further improve the full ensemble performance. }


\subsection{More Relevant Experiments} 
\label{sec: more experiments}
In this section, we show more related experiments. 
We perform classification on the much smaller ``invalid order'' nonsense subset (Section~\ref{sec:invalid order}) and show that \ModelName still works for this  low-resource setting, and doesn't require more fine-grained annotations like ``invalid order (self)'' and ``invalid order (other)''. 
In Section~\ref{sec:wrong justification}, we use \ModelName to generate discriminative replies  for categories like ``wrong justification'' that are hard to manually design replies for.

\subsubsection{Low-resource Setting on ``Invalid Order''}
\label{sec:invalid order}
In this section, we focus on ``invalid order'' nonsense examples to explore \ModelName's ability under low-resource settings, and also see if we could get better results given more fine-grained nonsense type annotations. 
Invalid order is an important category of nonsense, but we only have 79 annotated invalid orders in the train set, 9 in the validation set and 14 in test. It can be further split into two more fine-grained categories: proposing invalid orders for other players (other-invalid, 48 training examples), and proposing invalid orders for the player themselves (self-invalid, 33 training examples).

We use \ModelName to generate replies using the 79 annotated invalid orders  and the original 561 good situations, and  
Table~\ref{tab:invalid order} shows the classification results. The hand-crafted replies achieve a test F1 of 15.38, while \ModelName achieves a test F1 of 27.27, better than the hand-crafted replies. This suggests that even with only 79 annotated bad examples, \ModelName is able to generate many discriminative replies to achieve good classification results. The generated response with the best performance is \emph{``that would not work as i''}. 

In all the previous experiments, we don't have fine-grained annotations for different nonsense types, but we are also curious about this question: if we do have more fine-grained nonsense annotations, and use \ModelName to generate replies for each fine-grained nonsense type (e.g., \emph{``I can't move there''} for ``other-invalid-order'' and \emph{``You can't move there''} for ``self-invalid-order''), could the more focused replies produce better classification results? To answer this question, we use \ModelName to generate replies for the two fine-grained categories  ``other invalid'' and ``self invalid'', and combine the generated replies for the classification. The combined results are also in the last row of Table~\ref{tab:invalid order}. 
The performance actually becomes worse than the case where we  lump the two categories together and then generate ($27.27$ VS $16.67$). Even if we control the number of replies, lumping the categories together is still slightly better ($17.39$ VS $16.67$). The major reason behind is that the number of bad situations matters, if we lump categories together, we will have more bad situations to generate the replies and thus we can generate more better replies, which leads to a better classification performance. 

\subsubsection{Hard Categories without Hand-crafted Replies}
\label{sec:wrong justification}

In \ModelName, we tune and select the parameters and prune metrics that keeps the most hand-crafted responses, but there are categories we don't know how to design manual replies, such as ``wrong justification'', how do we tune the parameters? For these hard categories, we could try to simply use $\Delta_r^*=\min(\log P_{B}(r)) - \max(\log P_{G}(r)) > t_{\Delta}=0$ as the prune metrics, to generate replies that can completely separate the nonsense situations and good situations. Intuitively, 
it means that $r$ can completely separate the good examples and bad examples using its probability, and it can be used directly without tuning the threshold.

In this section, we focus on the subset of ``wrong justification'' $B_{w} = \{B_{i, w}\}$. The train set contains 40 ``wrong justifcation'' examples and 4149 good examples. The test set contains 11 ``wrong justifcation'' examples and 518 good examples. 

Because there are too few ``wrong justification'' examples in the train compared to the good examples and $\Delta_r^*$ is a strict metric, directly using $\Delta_r^*$ and the 561 random good distractors as contrastive examples leads to no generated replies. The ideal solution would be to ask human experts to make minimal edit to the original nonsensical message to make it reasonable, but this is too costly. 
So we estimate this process and use a newer language model $\mathcal{L'}$ trained on more data to generate a new message replacing the original nonsensical message, assuming that $\mathcal{L'}$ is better than the original $\mathcal{L}$ and thus will generate less nonsense. 
In this way, each ``wrong justification'' example $B_{i, w}$ has one contrastive example $G_{i, w}'$ that corresponds to it. $B_{i, w}$ and $G_{i, w}'$ differ only in the last message. 
We call this generated set of good examples  $G_{w}' = \{G_{i, w}'\}$. 

Contrasting $B_{w}$ with $G_{w}'$, we use \ModelName and $\Delta_r^*$, and generate 5486 replies. Because this category contains very limited examples (only 7 bad examples in valid, and 11 bad examples in test), the validation performance is not representative of the test performance. Instead of using the validation set to pick the replies, we directly list the top replies with the best test performance in Table~\ref{tab:wrong justification}. Note that this result is for reference only and not representative, because in practice, we cannot pick the best reply based on the performance on the test set. But we do show that \ModelName can find replies that work for the test set, and future research should focus on under few-shot settings, how to more effectively utilize the validation set to identify the best replies that would work for the test set.  
We also show the test performance of ``\ModelName (num=2805)'' on this subset for comparison (since it is the best model on the whole set, and uses hand-crafted replies to tune the parameters). The best reply is ``hmm, thats the way'' with a test F1 of 0.3077. Because $G_{w}'$ is an estimation of contrastive examples, and the rewritten messages are not necessarily about ``correct justification'' (could be about order proposal, or anything). 
So \ModelName-generated replies might capture the semantic meaning of ``justification'' (because it's contrastive to order proposal, etc), instead of ``correct'' justification. 
That's why the top replies also include ``yes i see your point,'', which is usually a follow-up reply for making justification. But from the test F1 comparable to the best \ModelName model on the whole set, we see that \ModelName is still doing its job in discriminating the bad situations and the good situations. And we believe that if the contrastive examples are related to ``correct justification'', \ModelName is able to generate more human-understandable wrong-justification-related replies.

If we reduce the training examples further to only five examples in $B_{w}$ and the five good examples correspondent to them, we can still generate 9673 replies (because there are only five good examples to contrastive against, we prune less and obtain more replies), the top ones are also listed in Table~\ref{tab:wrong justification}. This shows with a proper contrastive example set,  even if the bad situation is hard to design manual reply for and even if we only have super limited annotations, \ModelName can still generate large number of discriminative replies.

\begin{table}[!htbp]
    \centering
    \begin{adjustbox}{width=0.95\columnwidth}
    \begin{tabular}{l|l|ccc}
    \toprule
    \multirow{2}{30mm}{\textbf{Model}} &\textbf{Best Reply}& \multicolumn{3}{c}{\textbf{Test (518/11)}} \\ 
    \cmidrule(lr){3-5} 
    &&  \textbf{Prec} & \textbf{Recall} & \textbf{F1}  \\
    \toprule

    
    \multirow{5}{30mm}{\ModelName (num=2805) 
    on ($B$, $G$)}    &that is an excellent idea actually&20.59&50.00&29.17 \\

& ok, but if that fails&21.37& 35.71& 26.74\\ 
    
    &i will take it in&26.09& 25.71& 25.90 \\
    &ok, but if that works&22.22& 28.57& 25.00\\
    &that is not how this website&50.00& 15.71& 23.91 \\
    \midrule
    \multirow{5}{30mm}{\ModelName + $\Delta_r^*$ on  ($B_w$,  $G_w'$)} & 
    hmm, thats the way&26.67&36.36&30.77\\ 
    &well i already had&19.23&45.46&27.03\\ 
    &yeah that is actually a&16.67&63.64&26.42 \\
    &hmm, thats the&17.39&36.36&23.53\\
    &yes i see your point,&13.79&72.73&23.19\\ 
    \midrule
    
    \multirow{5}{30mm}{few shot, \ModelName + $\Delta_r^*$ on random 5 bad and good from ($B_w$,  $G_w'$)}&
    right, i had my&33.33&45.46& 38.46 \\
    &i see, i had my'&16.13& 45.46& 23.81\\
    &good point, i had my& 13.85& 81.82& 23.68\\
    &right, i had an&15.63& 45.46& 23.25 \\
    &right, i had my orders&12.35& 90.91& 21.74\\
    \bottomrule
    \end{tabular}
    \end{adjustbox}
    \caption{Single-reply classification result for the ``wrong justification'' subset, where we don't know how to design hand-crafted replies and thus cannot tune $\Delta_r$. So we use $\Delta_r^*$ directly to see if it's effective. The results show that directly using $\Delta_r^*$ and limited ``wrong justification'' examples achieves comparable performance with ``\ModelName (num=2805) on ($B$, $G$)'', which is tuned carefully and uses the full training set . }
    \label{tab:wrong justification}
\end{table}

\begin{table}[!htbp]
\centering
\small
\begin{adjustbox}{width=0.95\columnwidth}
\begin{tabular}{l}
\toprule
 \textbf{Hand-crafted Follow-up Reply}  \\
 \toprule
that move isn't possible\\
 that doesn't make sense\\
 are you talking about a different game\\
 i can't move to\\
 i have no units capable of doing that\\
 you can't do that\\
 what are you talking about\\
 i don't have units capable of doing that\\
 my army can't move there\\
 you said it twice\\
 you can't do that this turn\\
 i can't reach\\
 i guess you mean\\
 you just said that\\
 i don't understand\\
 you sent it to the wrong country\\
 you can't move to\\
 you are repeating yourself\\
 you are not bordering\\
 did you mean\\
 that's stupid\\
 i think you are hitting the refresh button\\
 your army can't move there\\
 i physically can't\\
 i am not bordering\\
 you can't move there\\
 you can't,\\
 i don't know what that means.\\
 you don't have any units bordering\\
 you don't have units capable of doing that\\
 you physically can't\\
 what do you mean?\\
 none of my units can do that\\
 i can't move there\\
 i don't have any units bordering\\
 i assume you mean\\
 you have no units capable of doing that\\
 stop repeating\\
 you don't have any units there\\
 you can't reach\\
 did you mean to send that to me?\\
 leave me alone\\
 none of your units can do that\\
 i am not able to move to\\
 that isn't a legal move\\
 i don't have any units there\\
 you aren't able to move to\\
\bottomrule

\end{tabular}
\end{adjustbox}
\caption{Hand-crafted follow-up replies.}
\label{tab:hand-crafted replies}
\end{table}

\begin{algorithm}[ht!]
\caption{\ModelName}
\small

\begin{algorithmic}[1]
\STATE \textbf{Input}: \\
Language Model $\mathcal{L}$, response prefix $r_{0}$, step $t$, \\
$p$, $K$, topn, prune step $t_{\text{prune}}$, max step $T$, 
bad messages examples $\{B_{\text{i}}\}$, good messages examples $\{G_{\text{i}}\}$

\STATE result = []

\STATE \# prune
\IF{$t$ >= $t_{\text{prune}}$ and need$\_$prune($r_{\text{cur}}$, $\{B_{\text{i}}\}$, $\{G_{\text{i}}\}$)}
    \STATE return []
\ENDIF

\IF{t < $T$}
    \STATE \# for each bad example, get next tokens to expand
    \STATE tok$\_$to$\_$bad$\_$exs = dict()
    \FOR{ex in $\{B_i\}$}
        \FOR{tok in get$\_$top$\_$tokens($M$, ex, $r_{\text{cur}}$, topp=p)}
            \STATE tok$\_$to$\_$bad$\_$exs.append(ex)
        \ENDFOR
    \ENDFOR
    \STATE \# for each good example, get next tokens to expand
    \STATE tok$\_$to$\_$good$\_$exs = dict()
    \FOR{ex in $\{G_i\}$}
        \FOR{tok in get$\_$top$\_$tokens($M$, ex, $r_{\text{cur}}$, topp=p)}
            \STATE tok$\_$to$\_$good$\_$exs.append(ex)
        \ENDFOR
    \ENDFOR

    \STATE \# sort based on the number of bad examples
    \STATE tok$\_$to$\_$bad$\_$exs = sorted(tok$\_$to$\_$bad$\_$exs)
    \STATE tok$\_$to$\_$bad$\_$exs = tok$\_$to$\_$bad$\_$exs[:topn]

\FOR{tok in tok$\_$to$\_$bad$\_$exs}
\STATE bad$\_$exs$\_$for$\_$token = tok$\_$to$\_$bad$\_$exs[tok]
\STATE good$\_$exs$\_$for$\_$token = tok$\_$to$\_$good$\_$exs[tok]
\IF{bad$\_$exs$\_$for$\_$token>=k}
    \STATE result += \ModelName($M$, $r_{\text{cur}}$+tok, t+1, p, k, topn, bad$\_$exs$\_$for$\_$token, good$\_$exs$\_$for$\_$token)
\ENDIF
\ENDFOR

\STATE \# keep the discriminative responses only
\IF{discriminative$\_$enough($r_{\text{cur}}$+tok,  bad$\_$exs$\_$for$\_$token, good$\_$exs$\_$for$\_$token)}
        \STATE result += [$r_{\text{cur}}$]
    \ENDIF

\ENDIF
\end{algorithmic}
\label{algorithm}
\end{algorithm}

\section{Brief Description of Diplomacy}
\label{appendix:rules}

Diplomacy is a seven player board game, where each player controls one of the seven European powers (\emph{Austria, England, France, Germany, Italy, Russia, and Turkey}) and competes to control the majority of \emph{supply centres} in Europe starting in the year 1901. The board is a map of Europe that is divided into 75 regions (split across land, water and coastal regions), 34 of which are \emph{supply centres (SCs)}. There are two types of units in the game: fleets, that can occupy water and coastal regions and armies, that can occupy land and coastal regions. Every power starts with 3 units and control 3 SCs, while Russia starts with 4 units and 4 SCs. As players control more SCs, they can build new units and when they lose SCs, they have to disband them. 

The game is split into years and each year contains multiple phases. The players privately issue commands for each unit they own at the end of each phase and these orders are then simultaneously revealed. Between phases, the players are allowed to communicate with other players in order to negotiate and coordinate their actions. This aspect is crucial, as the game is specifically designed so that a player is unlikely to achieve victory without the support from other players. A player wins the game by controlling 18 SCs. However, a game may also end in draw on any turn if all remaining players agree. When a player is in the lead, it is common for the remaining players to cooperate in order to prevent that player from winning the game and forcing a draw. See \citep{paquette2019no, kuliukas} for a more detailed description of the game.

\end{document}